%% file: main.tex
\documentclass{article}

\usepackage[nonatbib,preprint]{neurips_2020}

\usepackage[utf8]{inputenc} % allow utf-8 input
\usepackage[T1]{fontenc}    % use 8-bit T1 fonts
\usepackage{hyperref}       % hyperlinks
\usepackage{url}            % simple URL typesetting
\usepackage{booktabs}       % professional-quality tables
\usepackage{amsfonts}       % blackboard math symbols
\usepackage{nicefrac}       % compact symbols for 1/2, etc.
\usepackage{microtype}      % microtypography

\usepackage{amssymb}
\usepackage{amsmath}
\usepackage{latexsym}
\usepackage{color}
\usepackage{graphics}
\usepackage{xspace}
\usepackage{graphicx}  %Required
\usepackage{epsfig}
\usepackage{dsfont}
\usepackage{algorithm}
\usepackage[noend]{algpseudocode}
\usepackage{amsfonts}
\usepackage{url}
\usepackage{multirow}
\usepackage{subcaption}
\usepackage{enumitem}
\usepackage{mdwlist} % always use this to get tighter lists like enumerate and itemize
\usepackage{wrapfig}
\usepackage{blindtext}
\usepackage{pifont}

\newcommand{\ourmethod}{\textsf{NetReAct}\xspace}
\newcommand{\ourmethodviz}{\textsf{NetReAct-Viz}\xspace}
\newcommand{\netgist}{\textsf{NetGist}\xspace}
\newcommand{\nofeedback}{\textsf{FeedbackFree}\xspace}
\newcommand{\starspire}{\textsf{StarSPIRE}\xspace}
\newcommand{\spectral}{\textsf{Spectral}\xspace}
\newcommand{\metric}{\textsf{Metric-Learning}\xspace}
\newcommand{\community}{\textsf{Community-Det}\xspace}
\newcommand{\coarse}{\textsf{CoarseNet}\xspace}

\newcommand{\vast}{\textsc{VAST 2007}\xspace}
\newcommand{\crescent}{\textsc{Crescent}\xspace}

\newcommand{\n}{NB}

\newcommand{\hideContent}[1]{}

\newcommand{\Qprob}{F_{prob}}
\newcommand{\Qvalue}{\textrm{Q-value}\xspace}
\newcommand{\fbvalue}{\textsf{Feedback-value}\xspace}
\newcommand{\myurl}{\url{http://bit.ly/2MJzpWy}}

\newtheorem{definition}{Definition}

\title{NetReAct: Interactive Learning for Network Summarization}

\author{
    Sorour E. Amiri\textsuperscript{\rm *} \\
    \texttt{esorour@vt.edu}\\
    Virginia Tech
    \And
    Bijaya Adhikari\textsuperscript{\rm $\dagger$} \\
    \texttt{bijaya-adhikari@uiowa.edu}\\
    University of Iowa
    \And
    John Wenskovitch\textsuperscript{\rm $\ddagger$} \\
    \texttt{john.wenskovitch@pnnl.gov}\\
    PNNL
    \And
    Alexander Rogriguez\textsuperscript{\rm $+$} \\
    \texttt{arodriguezc@cc.gatech.edu} \\
    Georgia Tech
    \And
    Michelle Dowling\textsuperscript{\rm $\diamond$}\\
    \texttt{dowlinmi@gvsu.edu}\\
    GVSU
    \And
    Chris North\textsuperscript{\rm *} \\
    \texttt{north@cs.vt.edu}\\
    Virginia Tech
    \And
    B. Aditya Prakash\textsuperscript{\rm $+$}\\
    \texttt{badityap@cc.gatech.edu}\\
    Georgia Tech
}

\begin{document}

\maketitle

\begin{abstract}
 Generating useful network summaries is a challenging and important problem with several applications like sensemaking, visualization, and compression. 
However, most of the current work in this space do not take human feedback into account while generating summaries. Consider an intelligence analysis scenario, where the analyst is exploring a similarity network between documents. 
 The analyst can express her agreement/disagreement with the visualization of the network summary via iterative feedback, e.g. closing or moving documents (``nodes'') together. How can we use this feedback to improve the network summary quality? 
 %There are several challenges towards this goal, as the human feedback can be inconsistent and is typically sparse (as the analysis task is usually exploratory, and the user herself only has a partial understanding of a final summary while starting out). 
  %in incorporating human feedback to generate high-quality network visualization, such as how to integrate human input with the original objective of the visualization? Sometimes human opinions can be inconsistent with the original objective of the visualization. How to deal with such situations? Also, human feedback is typically sparse in the real-world, especially as the analysis task is usually exploratory, and the user only has a partial understanding of a final visualization. How should we generalize sparse user feedback?
In this paper, we present \ourmethod, a novel interactive network summarization algorithm which supports the visualization of networks induced by text corpora to perform sensemaking. % that enables a more comprehensible visualization by a multilevel approach. 
\ourmethod incorporates human feedback with reinforcement learning to summarize and visualize document networks. %It first summarizes the network by grouping relevant nodes, and then lays them out in groups in which spatial proximity is mapped to group similarity. 
Using scenarios from two datasets, we show how \ourmethod is successful in generating high-quality summaries and visualizations that reveal hidden patterns better than other non-trivial baselines. %We also show it can detect related documents and group them better than the .
\end{abstract}

\section{Introduction}
\label{sec:intro}
\input{010intro}

\hideContent{
\section{Related Work}
\label{sec:related}
\input{060related}

}

\section{Proposed Method}
\label{sec:proposed_method}
\input{040proposed}

\section{Empirical Studies}
\label{sec:experiments}
\input{050experiments}

\section{Conclusions and Discussion}
\label{sec:conclusions}
\input{070conclusion}

{
\small
\par\noindent\textbf{Acknowledgements}
This paper is based on work partially supported by the NSF (Expeditions CCF-1918770, CAREER IIS-2028586, RAPID IIS-2027862, Medium IIS-1955883, NRT DGE-1545362), NEH (HG-229283-15), CDC MInD program, ORNL, funds/computing resources from Georgia Tech and a Facebook faculty gift.
}

\clearpage
\newpage

\input{main.bbl}
\end{document}

%% file: 010intro.tex
%Understanding large-scale structured and unstructured document corpora has many applications such as social media analysis, security, marketing, and anomaly detection. Visualizing such datasets in a two-dimensional space is a common way of exploring them and understating their characteristics~\cite{munzner2014visualization}. However, doing this for a large and complex document corpus is often challenging, and is sometimes an impossible task to do meaningfully on a medium-size screen such as regular desktop monitors. Using human-in-the-loop techniques to develop an interactive visualization can help in understanding and visualizing such datasets more accurately on such screens. 

Networks occur in various domains such as social networks, entity networks, communication networks, population contact networks, and more. A meaningful summary of these networks can help users with various downstream tasks like sensemaking, compression, and visualization~\cite{adhikari2018propagation,amiri2018efficiently,purohit2014fast,shah2015timecrunch}. However, most prior work focus on generating summaries without human input~\cite{karypis:metis:sc98,purohit2014fast}.
In contrast, there are several applications, especially exploratory tasks, where incorporating human feedback in the summarization process is essential for generating useful summaries. 
%This can helps generate network summaries which closely aligns with the end application for which the summary is leveraged. 
For example, consider intelligence analysis~\cite{bradel2014multi}, which often involves making sense of networks of unstructured documents (like field reports) and extracting hidden information (like a terrorist plot) from a small subset of documents in the corpus. Users can provide feedback by interacting directly with the data, providing \textit{semantic interactions}~\cite{DBLP:journals/tvcg/EndertFN12} such as moving two nodes (documents) closer to each other to express similarity. This feedback helps the system to determine the relative importance of other documents with respect to the interests of the user.

Motivated by above, in this paper we tackle the novel problem of \textit{learning} to generate \textit{interactive} network summaries that incorporate user feedback. We showcase the usefulness of our summaries by focusing on an exploratory document visualization task. We leverage our summary to generate network visualizations, with the goal of aiding in investigating document networks and supporting the human sensemaking process; i.e., help the users ``connect the dots'' across different documents and discover the hidden stories behind them. More specifically, we try to answer: Given a document corpus represented as a network, can we \textit{learn} a model that incorporates user feedback alongside the objectives of the analysis task to generate high quality summary? Additionally, can we then leverage the summary to generate a meaningful visualization, and can such a model be re-applied to other document corpora?
%In this paper, we tackle the problem of visualizing document corpora to help the user `connect the dots' across different documents and discover the hidden stories behind them. More specifically, we try to answer the following questions:
% To design such as system, we need to answer the following questions: Given a document corpus and an analysis task, can we \textit{learn} a model that incorporates user feedback alongside the objectives of the analysis task from first principles? Can such a system be re-applied to other document corpora once it has learned how to visualize them? Can we leverage reinforcement learning (RL) to gradually update the model based on the received feedback from the user? Can this learned model help us to identify which documents are similar to each other and \emph{relevant} to the hidden story behind the documents? Can we use a feedback-based RL approach to summarize the document corpus by grouping related documents and identify the relationships between these groups to highlight the relevant documents to the given task? Can we subsequently generate a high-quality visualization based on the learned model and summarized corpus?
% Broadly, we study the novel problem of learning a network-based, interactive visualization with the aim of investigating text data to answer these questions.
Towards solving this problem, we face two major challenges. The first is \textit{simplicity of feedback.} Generally, the users are not experts in the summarization/visualization models and can only provide high-level semantic interactions. % We should be able to interpret their high-level semantic interaction into low-level instructions for the summarization and visualization models.%Generally, the users providing the feedback are not experts in the summarization/visualization models and cannot directly interact with the underlying model of the system. % We should be able to interpret their high-level semantic interaction into low-level instructions for the summarization and visualization models. 
The second challenge is \textit{sparsity and inconsistency of human feedback.} Getting human feedback is a slow and expensive process as the user needs to understand both the data and task in hand. As a result, the feedback is sparse.

Here the \emph{network} data structure represents a document corpus. Each node in the network represents a document, and  edge weights indicate the similarity between the documents based on their word usage. Our goal is to generate a summary network by grouping similar nodes (i.e., relevant documents together) and  finding hierarchical ``super-nodes'' (representing collection of documents) and ``super-edges'' (representing similarities between groups of documents). We then visualize this network summary to produce a high-level understanding of the entire corpus and highlight hidden patterns. We will discuss the concepts of super-nodes and super-edges later in detail.

%We then try to simplify the entire network down to a significant story-network and to organize it in a logical way on the screen. One can propose to simplify the network representation of text corpora by grouping the similar/relevant documents together to produce a high-level understanding of the entire corpus and highlight hidden patterns. Hence, we can model this approach as a network summarization problem, which consists of finding hierarchical ``super-nodes'' (representing collection of documents) and ``super-edges'' (representing similarities between groups of documents) and laying out the resulting smaller network based on the relatedness of the super-nodes/edges. Later on we discuss the concepts of super-nodes and super-edges in detail.

Our main idea is to tackle the above challenges via a reinforcement learning (RL) approach~\cite{watkins1992q} to summarize networks. We believe RL is especially suited for this problem, as it makes it possible to re-apply the learned model in similar scenarios to reduce the necessary amount of human feedback to gain useful information. We design \ourmethod, a feedback-based RL algorithm which integrates user interests with the objective of the summarization.  % Thus, this algorithm can also be used in visual analytics tools. 
%Some examples for user objectives are understanding the underlying story of the documents, detecting relevant documents to a given story, and highlighting the relationships between documents.
%\ourmethod learns a model to summarize a network of super-nodes and super-edges and gives a high-level understanding of the entire text data.
% \ourmethod also makes it possible to re-apply the learned model in similar scenarios to reduce the necessary amount of human feedback to gain useful information. 
\ourmethod also provides a multi-level understanding of the data by generating summaries of various sizes, enabling the user to zoom in to each document group to view these different summaries. %Figure \ref{fig:teaser} presents an example summary network and visualization generated by our approach.

The main contributions of this paper are:
\begin{itemize*}
    \item \textit{Incorporating human feedback.} We introduce \ourmethod, a novel approach which leverages feedback-based reinforcement learning to principally incorporate human feedback to generate meaningful network summaries. %\ourmethod makes it possible to re-apply the learned model to similar situations to reduce the amount of required feedback to generate a high-quality summary. \ourmethod is tested on real-world networks with various sizes to meaningfully visualize them for real-world applications.
    
    \item \textit{Meaningful relationships between groups as a summary network.} \ourmethod not only groups relevant nodes into super-nodes, but it also defines relationships between super-nodes. The weight of the edges in the summary network by \ourmethod represent the similarity between groups. %This helps the user to explore the corpus effectively by navigating through different super-nodes.
       
    \item \textit{Multi-scale visualization.} We leveraged \ourmethod to develop a multi-scale, interactive document network visualization framework. This visualization groups documents to summarize the network hierarchically, which  makes it possible for the user to get a multilevel understanding of the document network by looking at the summaries on different levels. %Depending on the task, users can choose how much detail they want to see about the dataset. %Also, by only investigating documents in one of the super-nodes, users can locally study the data in more detail. 
\end{itemize*}

%The rest of the paper is organized as follows:  first, we discuss related work in Sec.~\ref{sec:related}. Following this, we describe \ourmethod in Sec.~\ref{sec:proposed_method} in detail. We then discuss empirical results of our summarization approach on many datasets in Sec.~\ref{sec:experiments}, finally concluding in Sec.~\ref{sec:conclusions}.

%% file: 060related.tex
Our work is related to network summarization, graph drawing and document visualization, interactive reinforcement learning, and deep learning for graphs. We briefly discuss these areas here.  In contrast with the work that follows, we present an interactive visualization framework that incorporates human feedback with deep reinforcement learning.

\par \noindent \textbf{Network Summarization.}
There has been significant research interest in summarizing networks~\cite{liu2016graph}. Purohit et al.~\cite{purohit2014fast} proposed graph coarsening approach to preserve the diffusive characteristics of the network. Similarly, Karypis et al. proposed network summarization approach to preserve the communities in the network. There also has been several works for summarizing networks with respect to their temporal patterns~\cite{adhikari2018propagation,shah2015timecrunch,DBLP:conf/aaai/AmiriCP17}, network attributes~\cite{amiri2018efficiently,qu2014interestingness}, and so on. It is also related to the graph sparsification problem~\cite{mathioudakis2011sparsification}. \netgist~\cite{amirinetgist} has been recently proposed for task-based summarization of graphs using deep learning, but it handles only a restricted class of tasks, has not been designed for visualization, and cannot handle online human feedback. Unlike previous approaches, \ourmethod \emph{learns} network summaries automatically for interactive network visualization while satisfying online human feedback.

\par \noindent \textbf{Graph drawing and text analytics.}
Researchers have extensively studied the problem of drawing and visualizing graphs  \cite{herman2000graph}. Many of the approaches heavily rely on user input through direct manipulation of a graph. 
For example, ForceSPIRE~\cite{DBLP:journals/tvcg/EndertFN12} and \starspire\cite{bradel2014multi,Andrews:2010:STL:1753326.1753336} leverage the semantic interactions of users to visualize document networks.   Self et. al~\cite{DBLP:journals/tiis/SelfDWCWHLN18} propose an interactive metric learning method which enables parametric and
observation-level interaction to explore quantitative data.%, while Dis-Function supplements observation-level interaction with additional views to present facets of the data and dimensions~\cite{brown2012disfunction}.  
Similar semantic interaction techniques have demonstrated success in shielding users from the complexity of underlying models~\cite{endert2011oli}.  By manipulating the data in the projection rather than manipulating arcane model parameters, users are able to maintain focus on their analyses, thereby staying in the ``cognitive zone''~\cite{endert2013beyond,green2009building}.  %Work by Ruotsalo et al. uses these direct manipulation interactions to influence information retrieval~\cite{ruotsalo2013directing}, while similar work from Teevan et al. translates user feedback within a topic spatialization to tune search results incrementally~\cite{teevan2005personalizing}.  %VIGOR \cite{pienta2018vigor}  is an interactive visual analytics system for sense-making of graph query results in a cybersecurity setting. 
% with both top-down and bottom-up approaches. 
%FACETS~\cite{pienta2017facets}  helps non-expert users adaptively explore graphs and focus on neighborhoods that are most subjectively interesting. 
%: an analyst starts with a specific result and relaxes constraints or vis versa.
g-Miner~\cite{cao2015g} is an interactive system that enables visual mining of groups on multivariate graphs. There has not been much work on graph drawing from a data mining perspective~\cite{liu2016graph}.  %Typograph takes a multilevel abstraction approach, using extracted topics, keywords, and document snippets to assist in visualizing large text corpora~\cite{endert2013typograph}. 
Other popular works include constraint programming based force-directed layout for directed networks~\cite{dwyer2005dig}, simulated annealing based technique~\cite{davidson1996drawing}, and interactive clustering~\cite{DBLP:journals/tvcg/WenskovitchCRHL18, wenskovitch2017observation}.  %Most work in this line of research is task specific: hence these frameworks are not generalizable and cannot be used for other tasks.

\par \noindent \textbf{Information Synthesis}
A broad variety of information synthesis models exist within visual analytics systems, including network-based synthesis~\cite{mathams1988intelligence,stasko2008jigsaw}, spatial synthesis~\cite{kim2016interaxis,DBLP:journals/tvcg/EndertFN12}, entity profile synthesis~\cite{bier2006entity}, and interactive clustering~\cite{russell2006literate,wenskovitch2017observation,drucker2011helping}.  Most relevant to this work is spatial synthesis, which uses a space metaphor (often taking the form of a ``proximity$\approx$similarity'' relationship) to display similar documents and data points near each other, while dissimilar items are positioned at a distance.  Previous studies have demonstrated the utility of physical space to organize and synthesize text data~\cite{DBLP:journals/tvcg/EndertFN12,robinson2008collaborative,Andrews:2010:STL:1753326.1753336,andrews2012analystworkspace}. In this work, we leverage user-driven feedback to improve spatial synthesis using a cluster-based approach.

%%%%%%%%%%%%%%%%%%%%%%%%%%%%%%%%%%%%%%%%%%%%%%%%%%%%%%%%%%%
%%%%%%%%%%%%%%%%%%%%%%%%%%%%%%%%%%%%%%%%%%%%%%%%%%%%%%%%%%%
% Summarizing networks is a popular problem in many domains~\cite{liu2016graph} and has many applications in solving graph related tasks. For example, community detection problem \cite{karypis:metis:sc98}, influence maximization \cite{purohit2014fast} and compression \cite{navlakha2008graph}. Purohit et. al~\cite{purohit2014fast} propose a summarization technique for diffusion-related problems over networks. Other apporaches include subgraphs-based methods~\cite{koutra2014vog}, node and edge attributes-based methods~\cite{qu2014interestingness} and distant related field is network sparsification~\cite{mathioudakis2011sparsification}. Unlike previous approaches, We aim to \emph{learn} network summaries automatically for the task for interactive network visualization.

%%%%%%%%%%%%%%%%%%%%%%%%%%%%%%%%%%%%%%%%%%%%%%%%%%%%%%%%%%%
%%%%%%%%%%%%%%%%%%%%%%%%%%%%%%%%%%%%%%%%%%%%%%%%%%%%%%%%%%%

\par \noindent \textbf{Interactive reinforcement learning.} 
 The goal of interactive reinforcement learning is to incorporate human feedback to solve complex tasks in the real-world. 
 For example, Cruz et al.~\cite{7458195} leverages trainer advice to speed-up the learning process and increase the rate of success in the domestic task of cleaning a table. Some approaches~\cite{ng1999policy, tenorio2010dynamic, knox2008tamer, chang2006reinforcement} modify the reward function to integrate the human feedback with reinforcement algorithm. For example, Ng et al.~\cite{ng1999policy} explore the effect of altering rewards on optimal policy. Knox et al.~\cite{knox2008tamer} integrate human feedback by giving scalar reward signals in response to the agent's actions. Other approaches \cite{cederborg2015policy, riedmiller2018learning, knox2010combining, griffith2013policy, knox2012reinforcement} shape the policy directly based on human feedback. For instance, Griffith et al.~\cite{griffith2013policy} use human feedback as direct policy labels in a Bayesian approach and Cederborg et al.~\cite{cederborg2015policy} evaluate the policy-shaping algorithms using several human teachers.

%%%%%%%%%%%%%%%%%%%%%%%%%%%%%%%%%%%%%%%%%%%%%%%%%%%%%%%%%%%
%%%%%%%%%%%%%%%%%%%%%%%%%%%%%%%%%%%%%%%%%%%%%%%%%%%%%%%%%%%

\par \noindent \textbf{Deep learning for graphs.}
 There has been growing research work in recent years in leveraging deep learning for various graph mining tasks. Some use deep learning to embed nodes and networks~\cite{dai2016discriminative, grover2016node2vec}
 Others leverage deep learning and reinforcement learning approaches to propose meta-algorithms to solve combinatorial and optimization problems in graphs~\cite{amirinetgist, dai2017learning, bay2017approximating, nowak2017note, bello2016neural}. Also, researchers use deep learning approaches for cascade prediction~\cite{li2017deepcas}, 
 graph classification~\cite{niepert2016learning}, and graph kernels~\cite{yanardag2015deep}.

%% file: 040proposed.tex
In this paper, we focus on document network summarization to support visualization. Visualizing a document network helps users in their sensemaking process and understanding the documents, providing a two-dimensional illustration of the entire network and highlighting hidden patterns. Learning from the user is essential in designing an intelligent visualization framework that reflects the user's interests. Moreover, leveraging user feedback in the summarization helps to visualize data more effectively and efficiently than unsupervised approaches. Using supervised approaches are also not realistic in many real-world applications, particularly when the analyst is not an expert. %the field. 
We ground our work using a state-of-the-art systems for interactive document network visualization,   \starspire~\cite{bradel2014multi}. This system treats the document network as an attributed network, where each attribute is the frequency of a particular term in the document. It then builds a multilevel model of user interests based on the user's semantic interactions. 
In the \starspire framework, a user can interact with the system through semantic interactions such as minimizing documents, closing documents, annotating a document, highlight text within a document, and overlapping two documents. %\john{This list already exists 2 paragraphs up}. 
From these interactions, the models infer the importance of each attribute and then calculate the similarity between each pair of documents (nodes) based on weighted attribute similarity. However, since the number of unique terms is very high, the attributes are high-dimensional, and thus generating optimal weights requires a significant number of interactions.

In \ourmethod, we summarize document networks into explicit groups of related documents, demonstrating the relationship between groups in order to both generate a high-quality visualization for sensemaking tasks and to detect underlying stories hidden within the corpus.
%  We build our framework upon \netgist \cite{amirinetgist}, a task-based network summarization algorithm. 
We design \ourmethod to make it possible to incorporate semantic interaction feedback with network summarization, using that user guidance to generate a visualization of a document corpus. A ``good'' document network summary leads to a high-quality visualization, which helps a user to identify and read related documents and make sense of them quickly. More specifically, in a good network summary, each super-node (i.e., group) contains documents that are most relevant to each other according to the user's interest. Further, the structure of the network summary indicates the relationship between groups, which guides the user on how to navigate through different groups. Given such a summary, we first can visualize the summary network. After this, we can expand  the super-nodes that the user is interested in to suggest the most relevant documents. If the user wants to investigate more documents, we can then expand the closest (most similar) super-nodes to suggest another group of  relevant documents. 
%%%%%%%%%%%%%%%%%%%%%%%%%%%%%%%%%
%%%%%%%%%%%%%%%%%%%%%%%%%%%%%%%%%
%%%%%%%%%%%%%%%%%%%%%%%%%%%%%%%%%
\subsection{User Feedback}
\label{sec:userfeedback}

We observed several users completing a sensemaking task using the \starspire framework, and based on their behavior we selected a subset of the supported semantic interactions for generating user feedback. We divide such interactions into \emph{positive} and \emph{negative} feedback (see Table \ref{tab:feedback}). For example, positive feedback can indicate the user's intention to put two documents close to each other (i.e., group two nodes together), while negative feedback means they should be far from each other. Overlapping two documents indicates that the user agrees to display them close to each other. On the other hand, minimizing a document while reading another one is a sign of the disagreement with the visualization. This local feedback is then applied to the entire visualization and corpus. Such feedback is sparse, as the user cannot evaluate all documents and every aspect of the visualization.

\begin{table}[htp]
\centering
\caption{\ourmethod feedback types and corresponding semantic interactions in \starspire.}
\begin{tabular}{|l| p{2in}|}
     \hline
     \textbf{Feedback Type} & \textbf{Semantic Interaction} \\ \hline \hline
     Negative feedback&  (1) Minimizing document, (2) Closing document\\ \hline
     Positive feedback& (1) Annotation, (2) Highlighting document, (3) Overlapping document \\ \hline
\end{tabular}
\label{tab:feedback}
\end{table}
%%%%%%%%%%%%%%%%%%%%%%%%%%%%%%%%%
%%%%%%%%%%%%%%%%%%%%%%%%%%%%%%%%%
%%%%%%%%%%%%%%%%%%%%%%%%%%%%%%%%%
\subsection{Interactive Summarization Model}
In this section, we describe an interactive network summarization framework to incorporate the user feedback and address its sparsity.  Note that our goal is also to learn the steps to be taken for the summarization process so that the same approach can be re-applied on other document corpora with similar characteristics. 

%We show how such summarization helps in generating a high-quality visualization which reflects user interests and helps in revealing underlying patterns and giving an in-depth understanding of the dataset.
\vspace{2mm}
\par \noindent \textbf{Network Construction. }
We start by converting the given document corpus into a network $G(V, E, W)$, where nodes ($V$) represent documents and edges ($E$) and their weights ($W$) represent document similarity. We define the weight $w(v_1,v_2)$ to be the cosine similarity between the corresponding TF-IDF vectors of the documents. Note that $G$ is a complete-graph of size $|V|$.

%into a summary net
%by grouping nodes  similar/relevant document and generate a %summary graph $G^s(V^s, E^s, W^s)$. 
%We call such document summaries a summary network $G^s(V^s, E^s, W^s)$ where super-nodes ($V^s$) are the groups of related documents and super-edges ($E^s$) and their weights ($W^s$) show the similarity between them.

\vspace{2mm}
\par \noindent
\textbf{Network Summarization. } Once the network $G$ is constructed, the summarization process begins. The goal is to generate a smaller network $G^s(V^s, E^s, W^s)$ from the original network $G(V, E, W)$ such that nodes representing  similar/relevant documents in $G$ are grouped into a single node (a ``super-node'') in $G^s$.  Nodes in $G^s$ therefore represent a group of documents. We call $G^s(V^s, E^s, W^s)$ a ``summary network,'' where super-nodes ($V^s$) are the groups of related documents and super-edges ($E^s$), and their weights ($W^s$) represent the average similarity between group of documents represented by the two endpoints. We obtain $G^s$ via a series of ``assign'' operations on $G$. The ``assign'' operation assigns nodes to their super-nodes. This operation partitions the original network $G$ and groups each partition to form a super-node in the summary network $G^s$. Next, we must determine how to partition the original network $G$ in a meaningful manner. In other words, how to decide between two partitions of $G$, and how to measure the quality of each assigning operation and network summary? 

%Reinforcement Learning (RL)---more specifically Q-learning~\cite{sutton1998introduction}---is a natural solution to the above questions. In general, RL methods learn to take an ``\emph{action}'' in an ``\emph{environment}'' to maximize cumulative ``\emph{reward}'' based on a ``\emph{transition}'' function. 

Reinforcement Learning (RL) is a natural fit to answer the question above, as we can view our problem as taking actions (assigning nodes to a super-node) to maximize the reward (the final quality of grouping of documents). The next step in our summarization process is to formalize the RL framework for our task.

\subsubsection{Interactive Reinforcement Learning Formulation}
%Our goal is to use Q-learning which is known to be more sample efficient than policy gradient methods \cite{watkins1992q}.
We use Q-learning, as it is known to be more sample efficient than policy gradient methods~\cite{watkins1992q}.
Each RL framework has five main components: states, actions, transition function, reward, and policy. We further add an additional feedback component as we design an interactive RL formulation. Brief descriptions of each follow.
\vspace{2mm}
\par\noindent
\textbf{1. State:} The state $s$ is the sequence of actions which assign nodes to different super-nodes. We use the embedding $s = [l_1, l_2, \ldots, l_n]$, $\forall_{i \in \{1, 2, \ldots, n \}} \quad l_i \in \{1, 2, \ldots, |V_s| \}$ in  to represent the states in a vector of n-dimensional space. 
\hideContent{
\begin{align}
    \label{eq:rl_state}
    &s = [l_1, l_2, \ldots, l_n]\\
   & \forall_{i \in \{1, 2, \ldots, n \}} \quad l_i \in \{1, 2, \ldots, |V_s| \} \notag
\end{align}
}
\par\noindent
\textbf{2. Action:} 
An action $a$ at state $s_t$ selects a document $i$, assigns it to a new super-node $v^s$, and transfers it to the next state $s_{t+1}$. 
\par\noindent
\textbf{3. Transition:} We assume the transition function $\mathcal{T}$ is deterministic and corresponds to the action of assigning a document to a new super-node -- i.e., $\mathcal{T}(s_t, a) = s_{t+1}$.

\par\noindent
\textbf{4. Rewards:}
We define the reward to be $-1$ for a state $s$, unless it is a terminal state. A terminal state in our case is a state which satisfies all of the positive and negative feedback of the user. Intuitively the reward of $-1$ encourages the learner to satisfy the user feedback faster. Formally, we define our reward function as follows:
\begin{align}
\label{eq:reward}
    r(s,a) = \begin{cases}
\Qprob(s_{next})=\sum _{i=1}^{k}\frac{y_i^T A y_i}{y_i^T D y_i}  & \text{ if } s_{next} \text{ is a terminal state}\\ 
 -1& \text{otherwise }
\end{cases}
\end{align}
Here, $A$ is the adjacency matrix of the document graph $G$, $D$ is the diagonal matrix of node degrees, and $y_i$
is an indicator vector for super-node $v^s_i \in V^s$, i.e., $y_i (v) = 1$
if a node $v$ belongs to super-node $v^s_i$, zero otherwise. In Eq.~\ref{eq:reward}, we compute $\Qprob(s_{next})$, which measures the quality of the document groups. By maximizing $\Qprob(s_{next})$, we maximize the quality of document groups~\cite{DBLP:conf/sdm/WhangDG15}. 
\par\noindent
\textbf{5. Feedback}
We assume a case that the user is interacting with the system until she is happy with the visualization, a process of incrementally formalizing the space~\cite{shipman1999formality}. This means that we must learn the model until all the feedback from the user is satisfied. The feedback is in the form of positive and negative interactions (See Tab. \ref{tab:feedback}). The user can indicate if she agrees to group a pair of documents (i.e., positive feedback) or disagrees with it (i.e., negative feedback). 
We represent the feedback with two graphs, which we call feedback graphs. A positive feedback graph $G^+$ is created from the set of positive feedback, i.e., the edges in $G^+$ are pair of related documents that the user indicated. Similarly, the negative feedback graph $G^-$ is created from the set of negative feedback.

To satisfy all the feedback, we must group all positive feedback node pairs in the same super-nodes and all negative feedback node pairs in different super-nodes. These constraints can be stated using the positive and negative feedback graphs $G^+$ and $G^-$ as follows:
\begin{align}
\label{eq:feedback}
\sum_{i=1}^{k} y_i^T A_{G^+} y_i- \sum_{i=1}^{k}y_i^TA_{G^-}y_i = \sum A_{G^+} 
\end{align}
Here, $y_i$ is a super-node, $k$ is the number of super-nodes, and $A_G$ is the adjacency matrix of $G$. In the real world, we do not expect the user to provide all possible feedback, as this would essentially provide the desired summary without computational assistance. Rather, the provided feedback are sparse, especially when the task is exploratory in nature. To handle such problems, we combine the reward in Eq.~\ref{eq:reward} with feedback (Eq.~\ref{eq:feedback}). Thus, our goal is to achieve a summary that satisfies all feedback and maximizes the reward.
\par\noindent
\textbf{6. Policy:}
The policy function $\pi(a^*|s)$ specifies what action to take at a given state. It is defined as $\pi(a^*|s) = \arg\max \Qvalue(s,a)$, 
where $\Qvalue(s,a)$ is the Q-value of the state $s$ and action $a$ that estimates the expected cumulative reward achieved after taking action $a$ at state $s$. Our goal is to learn an optimal Q-value function resulting in the highest cumulative reward. We leverage the Q-learning, which iteratively updates $\Qvalue(s,a)$ until convergence~\cite{sutton1998introduction}.
\subsubsection{Q-learning}
\label{Sec:learn}
Our pipeline learns the best super-node for each node in the document graph $G$ such that its corresponding summary graph $G^s$ gives a high-quality visualization and is generalizable to similar unseen document corpora.  We use Q-learning to learn the pipeline. First, we define how to estimate the Q-value of a state $s$ and action $a$, $\Qvalue(s,a)$. We define the Q-value of a state and action as the expected rewards in the future as follows,
\begin{align}
\label{eq:bellman}
\Qvalue(s,a) = \mathbb{E}\left[ \sum_{t\geq0}\gamma^t r_t|s_0=s,a_0=a, \pi \right] 
\end{align}

Our aim is to find the maximum cumulative reward achievable with state $s$ and action $a$: $\Qvalue^*(s,a) = \max \Qvalue(s,a)$. We estimate $\Qvalue^*(s,a)$ iteratively using the Bellman equation:
\begin{align}
\Qvalue_{i+1}(s,a) = \mathbb{E}\left[ r + \gamma \max_{a'} \Qvalue_i (s',a')|s,a \right] 
\end{align}

We use a Fully Connected (FC) neural network to embed 
each state $s$ and get a compact representation of it. We combine the embedding layer with the $\Qvalue^*(s,a)$ estimator to have an end-to-end framework to summarize the document network.

In our framework, the input state $s$ is fed into the FC neural network. The output of this step is a compact representation of the state, which is then fed into another FC that decides how to update the super-nodes. Alg.~\ref{alg:summarization} presents an overview of our summarization algorithm.
\begin{algorithm}
\caption{Summarization}\label{alg:summarization}
{
\begin{algorithmic}[1]
\Require{$G$, $G^+$, $G^-$, $k$}
\State Randomly Initialize the deep Q-learning parameters
\State // learning how to summarize
\For{episode=1 \textbf{to}  $T$}
\State Initialize $s_0:$ Initialize $\{y_1, y_2,\ldots, y_k\}$
\While {$\fbvalue < k$}
\State // Take a action
\State $a^* = \arg\max\; \Qvalue(s_{current},a)$
\State $s_{current} = \mathcal{T}(s_{current}, a^*)$
\EndWhile
\State // Evaluate
\State Evaluate the corresponding partitioning to $s_{current}$ (Eq. \ref{eq:reward})
\State // Optimize (see Section \ref{Sec:learn})
\State Update deep Q-learning parameters for better summary 
\EndFor
\State \textbf{Return} the trained model and super-nodes $\{y_1, y_2,\ldots, y_k\}$ 
\end{algorithmic}
}
\end{algorithm}
\vspace{-0.2in}

\subsubsection{Hierarchical Summaries}
Our goal is not only to summarize the network but also to provide a multi-level understanding of the structure. This is specially useful in large networks, where it is challenging to meaningfully and efficiently generate the best summary. Hence, we propose a hierarchical approach, where we intuitively, zoom out one ``level'' at a time  to generate summaries with different sizes. Specifically, in each step \ourmethod tries to partition the data into two super-nodes, and then iteratively summarizes each part until reaching a summary with the desired size.

\hideContent{
Alg. \ref{alg:hierarchical} shows our hierarchical summarization approach, \ourmethod.  

\begin{algorithm}
\caption{\ourmethod}\label{alg:hierarchical}
{
\begin{algorithmic}[1]
\Require{$G$, $G^+$, $G^-$, $k$, $current_k$}
\State $y_1, y_2$ = Summarization($G$, $G^+$, $G^-$, $2$)
\If {$2 \times current_k \leq k$}
\State $G_1,G^+_1, G^-_1 \longleftarrow$ sub-graph \& feedback corresponding to $y_1$
\State $G_2,G^+_2, G^-_2 \longleftarrow$ sub-graph \& feedback corresponding to $y_2$
\State $y_{1,1}$,\ldots $y_{1,\frac{k}{2}}$ = Summarization($G_1,G^+_1, G^-_1$, $k$, $2\times current_k$)
\State $y_{2,1}$,\ldots $y_{2,\frac{k}{2}}$ = Summarization($G_1,G^+_1, G^-_1$, $k$, $2\times current_k$)
\State $y_1, \ldots y_k \longleftarrow $ combine $y_{1,1}$,\ldots $y_{1,\frac{k}{2}}$ and $y_{2,1}$,\ldots $y_{2,\frac{k}{2}}$
\EndIf
\State \textbf{Return} $y_1, \ldots y_k$
\end{algorithmic}
}
\end{algorithm}
}
\subsubsection{Generate Summaries}
\label{sec:summary}
After learning the best super-nodes of the network, we \emph{merge} nodes in the same super-nodes to generate a corresponding super-node. We also connect each super-node to others by super-edges. The weight of the super-edge from $y_1$ to $y_2$ is the average similarity between documents in $y_1$ to $y_2$. More formally, we define the merge operation as follows,

\begin{definition}
\label{def:merge}
 Merge operation merges nodes $v_1$, $v_2$, \ldots, $v_b$ into a new node $y$, such that $\forall_{j=1,\ldots , b}\;v_j\in V$. We add new edge $(y, i)$ for all the nodes $i \in \underset{j=1\ldots, b}{\bigcup} \n(v_j)$ with weight $\frac{\sum_{j=1}^{b}W(v_j, i)}{b}$. 
\end{definition}
We merge nodes in the same super-node using Def. \ref{def:merge} to yield the summary document network $G^s(V^s, E^s, W^s)$, where $|V^s| = k$.

\subsection{Two-step Visualization}
\label{sec:twostep}

Once the summary is generated, our goal is to visualize the document network. We design a multilevel framework that first leverages the weighted force-directed layout~\cite{dwyer2005dig} to visualize the summary graph. This gives us a 2D layout of the summary network, which we treat as the ``backbone'' of our visualization process. Note that each super-node consists of a group of nodes, which induce sub-graphs in the original network. We separately run the weighted force-directed layout on each sub-graph induced by the super-nodes. Finally, we ``combine'' the layouts within each super-node with the backbone layout of the entire summary network in a multi-level fashion to visualize with entire network. Lines 3-7 of Alg.~\ref{alg:ourmethod} show the pseudocode of this two-step visualization approach.

\begin{algorithm}
\caption{\ourmethodviz}\label{alg:ourmethod}
{
\begin{algorithmic}[1]
\Require{$G$, $G^+$, $G^-$, $k$, $current_k$}
\State $y_1, \ldots y_k =$  Hierarchical-Partitioning$(G, G^+, G^-,k, 1$)
\State $G^s \longleftarrow $ merge nodes in $y_1, \ldots y_k$
\State $\forall_{1 \leq i \leq k} loc_{y_i} \longleftarrow Layout(G^s)$
\For {super-node $y_i \in \{y_1,\ldots,y_k\}$}
\State $G_i \longleftarrow $ Corresponding sub-graph of $y_i$
\State $\forall_{v \in y_i}loc_v \longleftarrow Layout(G_i)$
\State $\forall_{v \in y_i}loc_v = \frac{loc_v}{k} + loc_{y+1}$
\EndFor
\State \textbf{Return} $\forall_{v \in V} loc_{v}$
\end{algorithmic}
}
\end{algorithm}

\vspace{-0.2in}

%% file: 050experiments.tex
%We design several experiments to evaluate the performance of \ourmethod and \ourmethodviz. 
\hideContent{
Specifically, we seek to answer the following questions:
\begin{itemize}
    \item \textbf{Q1.} Does \ourmethod learn to generate high-quality summaries?
    \item \textbf{Q2.} What is the effect of feedback on \ourmethod?
    \item \textbf{Q3.} Does \ourmethodviz give high-quality document visualizations?

\end{itemize}
}
We used Python and PyTorch to implement all steps of \ourmethod and \ourmethodviz, and our code is publicly available for academic and research purposes\footnote{\myurl}. %We implement deep Q-learning on GPUs, which makes it suitable for summarizing large networks.
%%%%%%%%%%%%%%%%%%%%%%%%%%%%%%%%%%%%%%%%%%%%
%%%%%%%%%%%%%%%%%%%%%%%%%%%%%%%%%%%%%%%%%%%%
%\subsection{Datasets and Baselines}
We explore the effectiveness of \ourmethod and \ourmethodviz on two document network datasets: \crescent~\cite{hughes2003discovery} is a document corpus containing synthetic intelligence reports related to terrorist activities, and the \vast Challenge dataset (``Blue Iguanodon'')~\cite{grinstein2007vast} contains documents regarding wildlife law enforcement subplots. We compare performance of \ourmethod against several baselines including \textbf{\spectral} \cite{von2007tutorial}, \textbf{\community} ~\cite{girvan2002community}, \textbf{\coarse} ~\cite{purohit2014fast} and \textbf{\metric} \cite{xing2003distance} based approaches.

\subsection{ Quality of Summaries}
\label{sec:quality_summary}
Here, we demonstrate that \ourmethod generates high quality network summaries by both grouping relevant nodes in the same super-node and satisfying the constraints posed by users as feedback. In this section, we focus on quantitative results. %We further analyze the summary networks qualitatively in Section~\ref{subsection:qualitativeAnalysis}.

We measure the quality of the summary by quantifying the ease of identifying relevant documents. To that end, we measure the purity of super-nodes that contain relevant documents. In other words, we calculate the average probability of observing a relevant document in a super-node that contains at least one relevant document. Formally,
\begin{align}
    \label{eq:purity}
    \rho = \frac{1}{|V^s_r|} \sum_{v^s \in V^s_r} Pr(doc = relevant|v^s)
\end{align}
where $V^s_r$ is the set of super-nodes that have at least one relevant document to the scenario, $v^s$ is a super-node in the set $V^s_r$ and $Pr(doc = relevant|v^s)$ is the probability that a document is relevant to the hidden scenario in $v^s$. Intuitively, if the value of $\rho$ is closer to one, it means the user can easily find relevant documents in a selected super-node.

\begin{figure*}[htp]
    \centering
    \begin{subfigure}[htb]{0.23\textwidth}
    \centering
    \includegraphics[width = \textwidth]{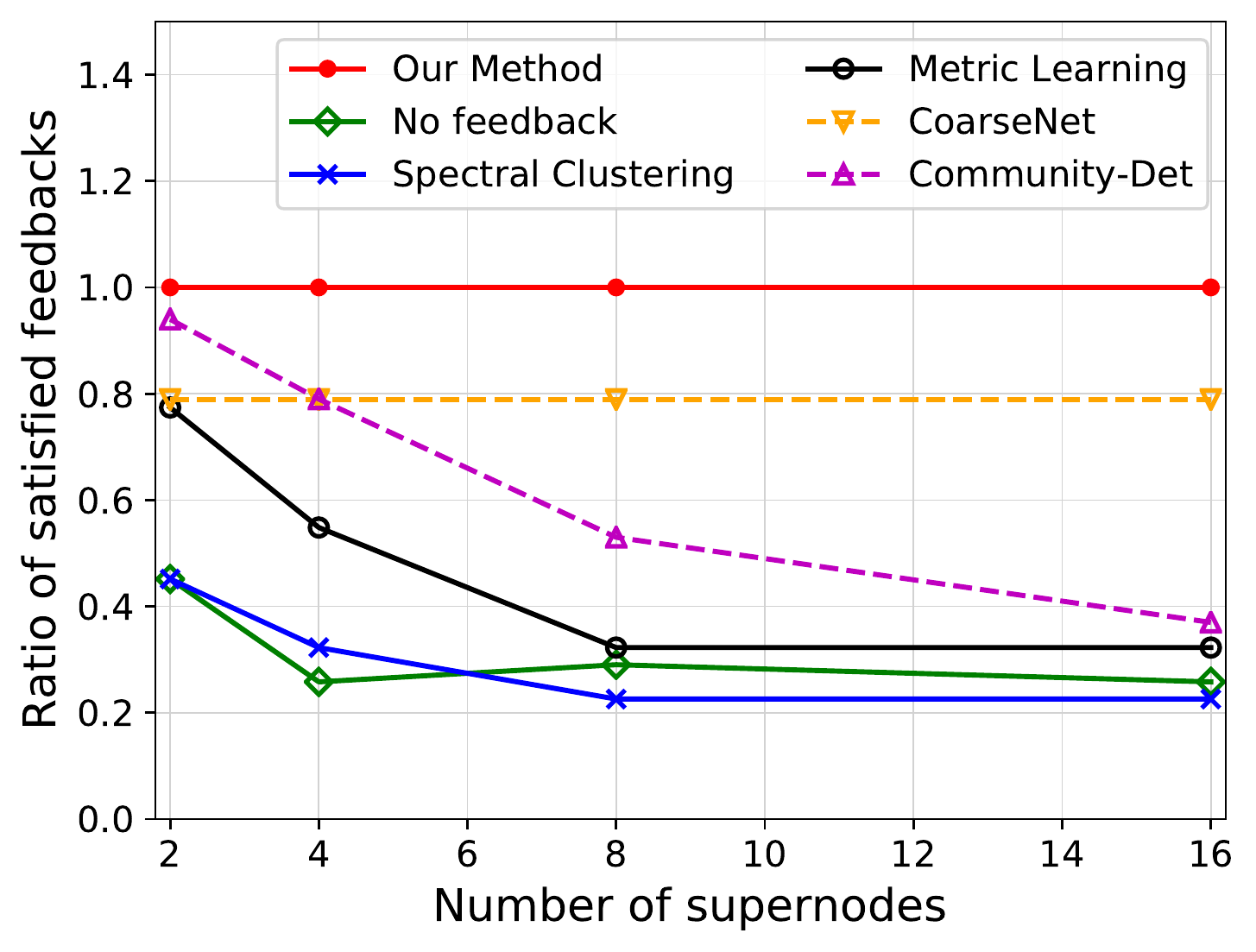}
    \caption{\crescent}
    \label{fig:crescent_nof_supernodes_satisfied}
    \end{subfigure}
    \begin{subfigure}[htb]{0.23\textwidth}
    \centering
    \includegraphics[width =\textwidth]{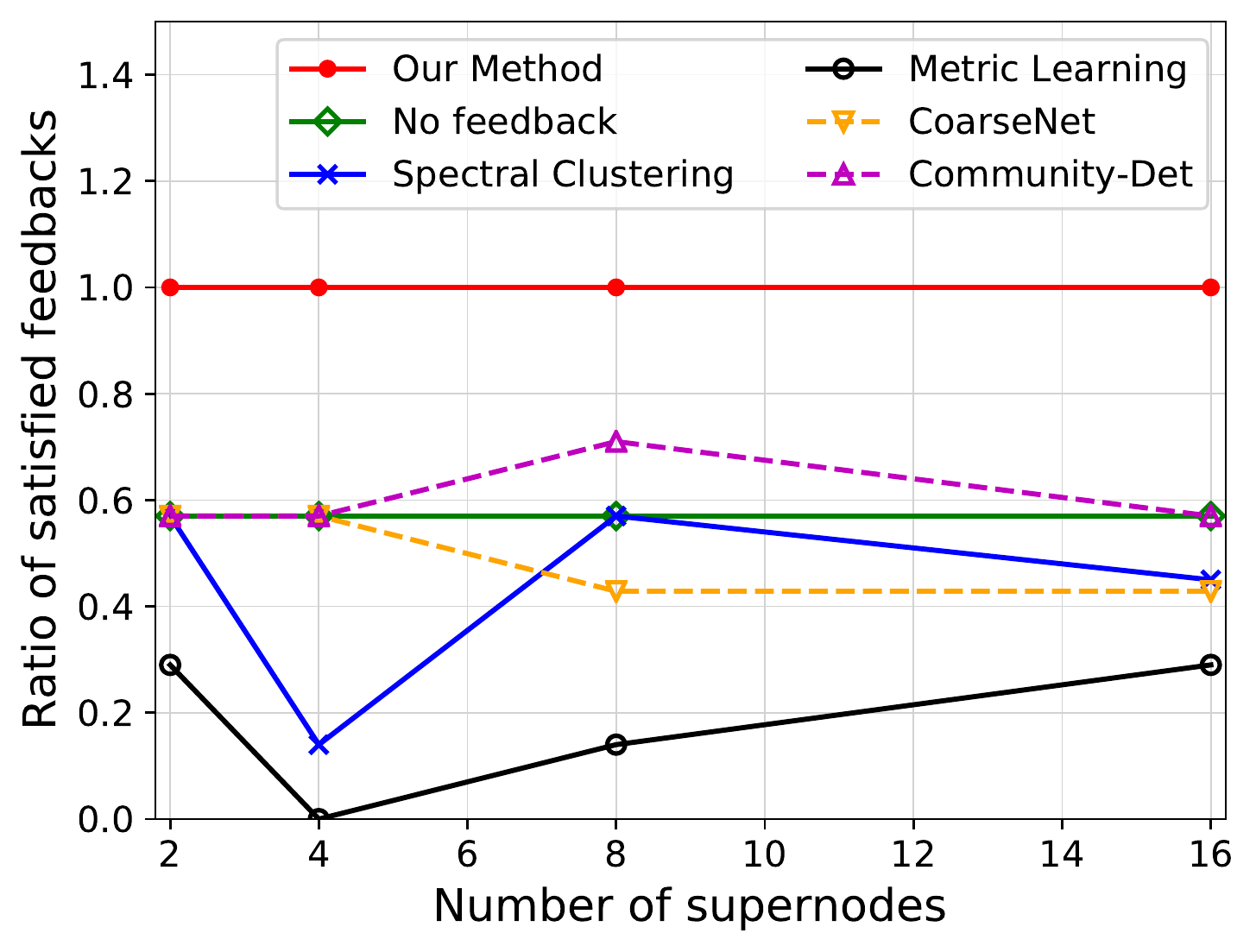}
    \caption{Chinchilla Bio-terror subplot}
    \label{fig:Chinchilla_Bioterror_VAST2007All_satisfied}
    \end{subfigure}
    \begin{subfigure}[htb]{0.23\textwidth}
    \centering
    \includegraphics[width =\textwidth]{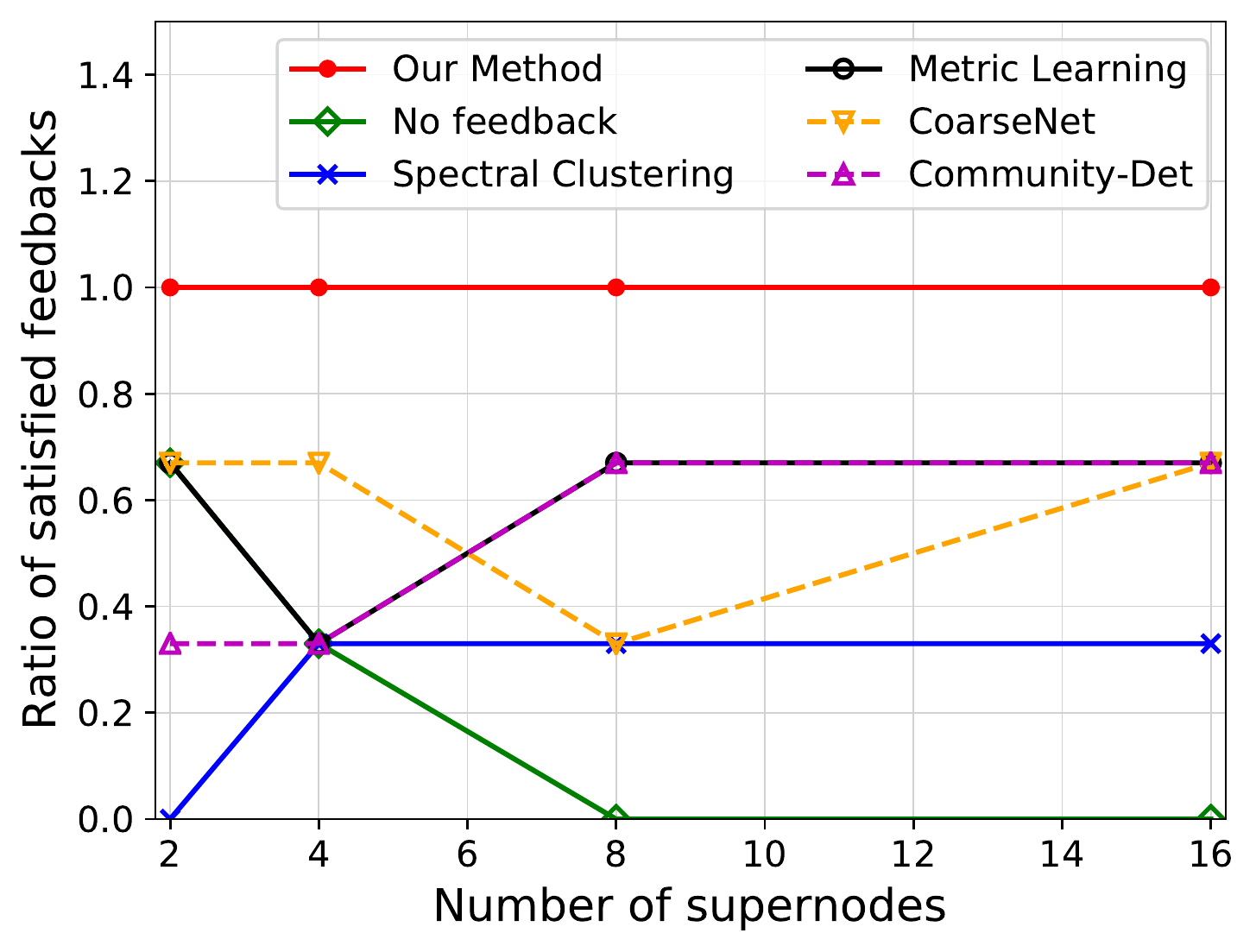}
    \caption{Bert subplot}
    \label{fig:Bert_Takes_a_Powder_VAST2007All_satisfied}
    \end{subfigure}
    % \begin{subfigure}[htb]{0.23\textwidth}
    % \includegraphics[width =\textwidth]{FIG/Fish_VAST2007All_satisfied.pdf}
    % \caption{Fish subplot}
    % \label{fig:Fish_VAST2007All_satisfied}
    % \end{subfigure}   
    \begin{subfigure}[htb]{0.23\textwidth}
    \centering
    \includegraphics[width =\textwidth]{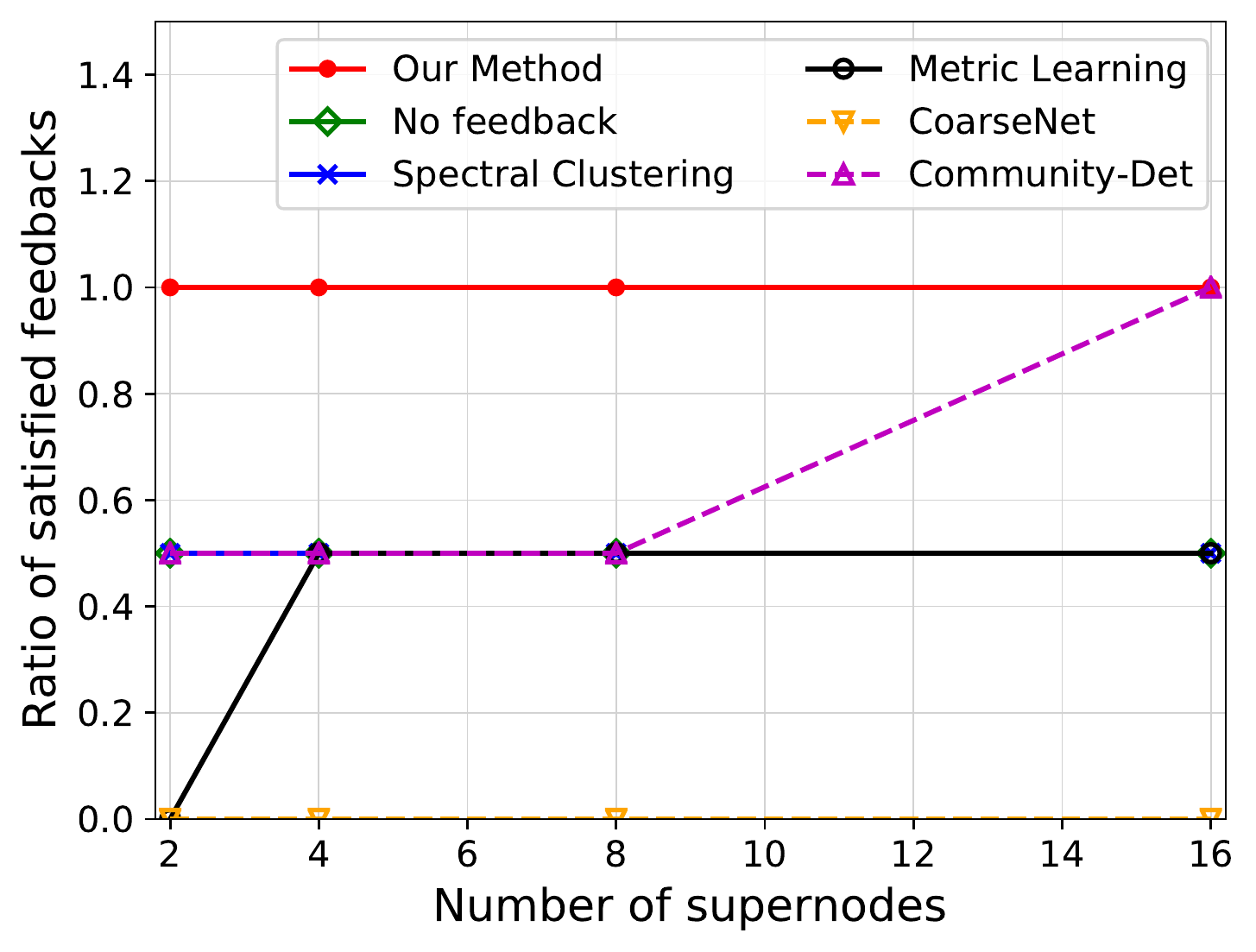}
    \caption{Circus subplot}
    \label{fig:Hassan_Circu_VAST2007All_satisfied}
    \end{subfigure}     
    \caption{Ratio of satisfied feedback in (a) \crescent and (b-d) different subplots of the \vast dataset. Note, \ourmethod satisfies all the user feedback while other baselines do not. }
    \label{fig:nof_supernodes_satisfied}
\end{figure*}

\begin{figure*}[htp]
    \centering
    \begin{subfigure}[htb]{0.23\textwidth}
    \centering
    \includegraphics[width=\textwidth]{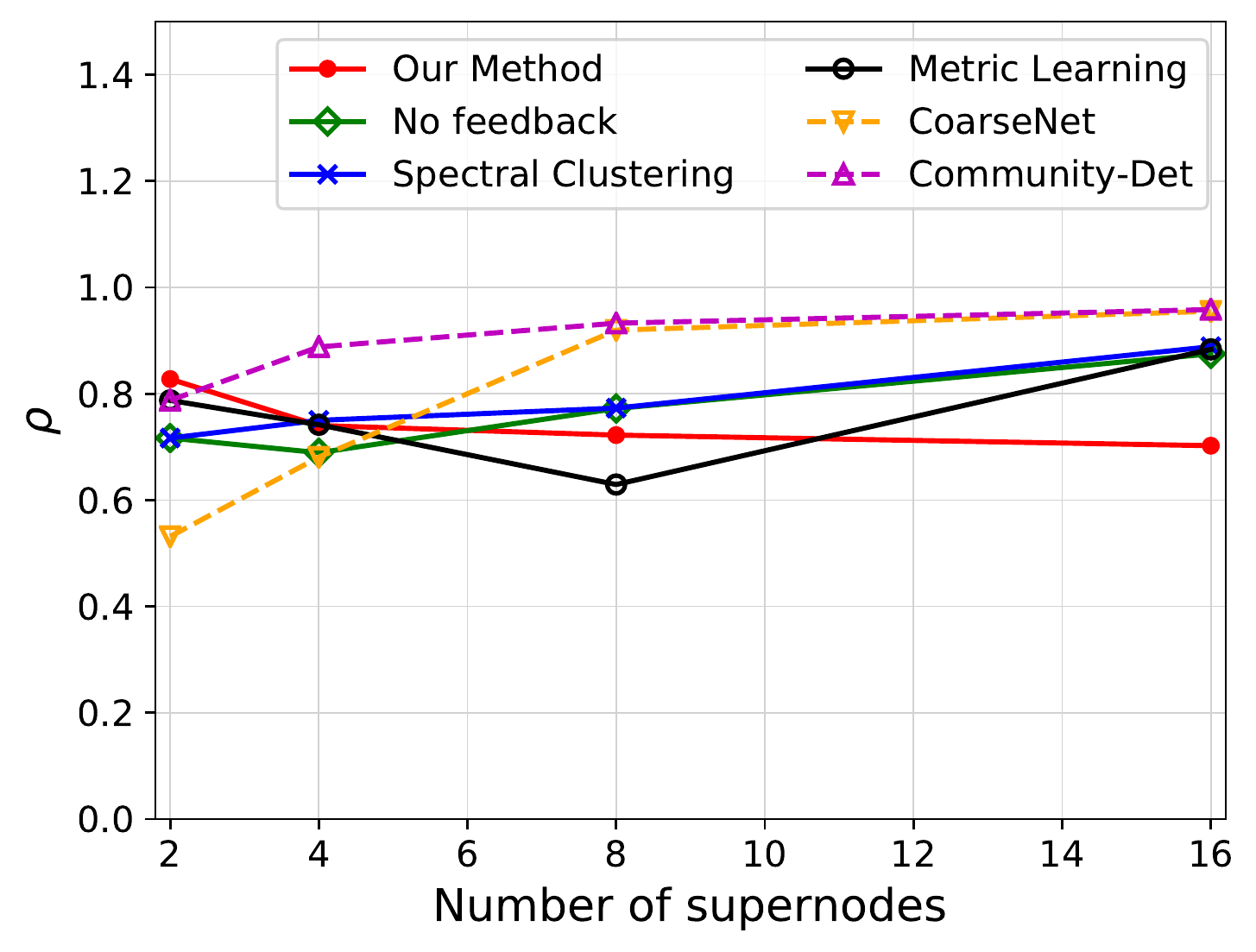}
    \caption{\crescent}
    \label{fig:crescent_nof_supernodes_rho}
    \end{subfigure}
    \begin{subfigure}[htb]{0.23\textwidth}
    \centering
    \includegraphics[width =\textwidth]{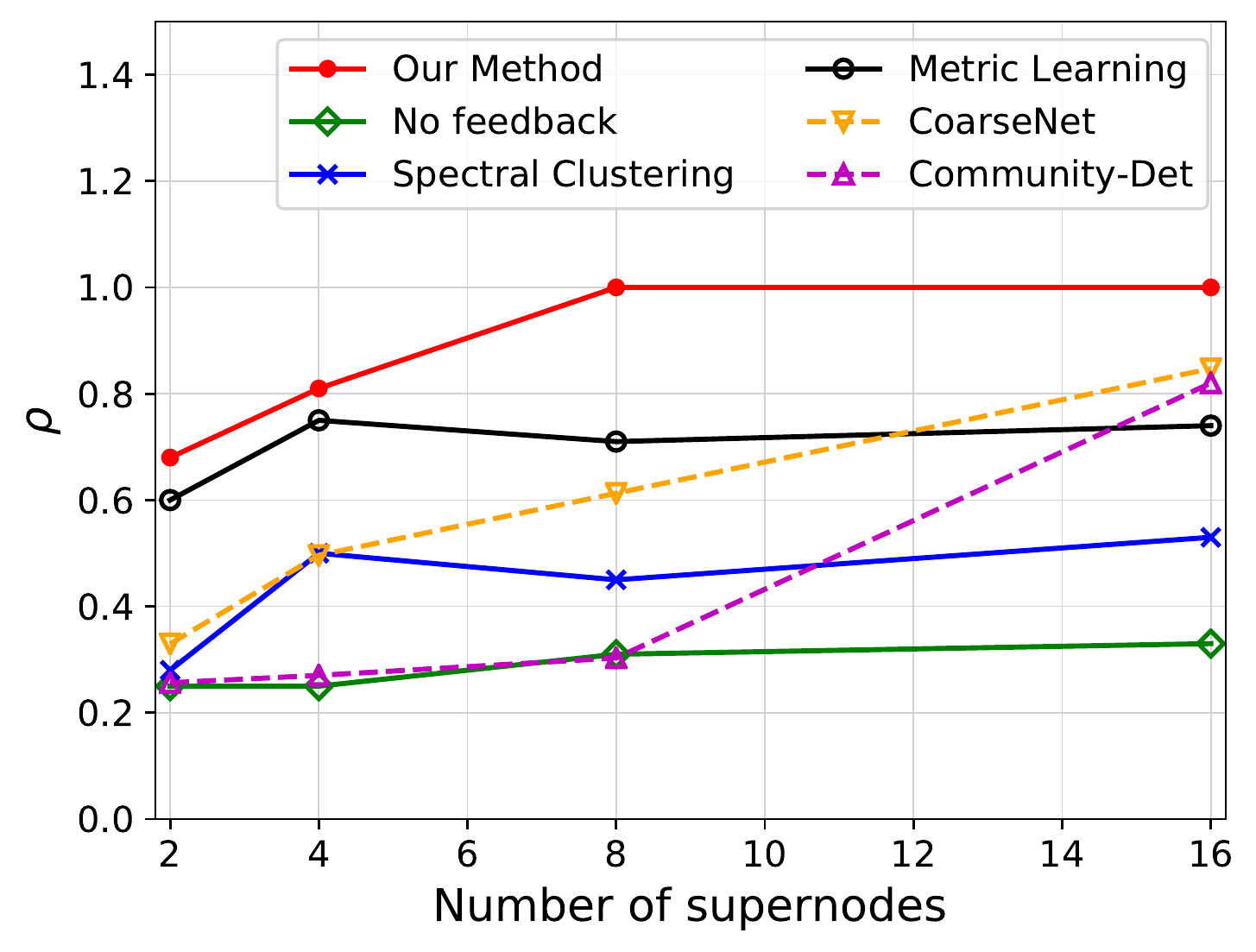}
    \caption{Chinchilla Bio-terror subplot}
    \label{fig:Chinchilla_Bioterror_VAST2007All_rho}
    \end{subfigure}
    \begin{subfigure}[htb]{0.23\textwidth}
    \centering
    \includegraphics[width =\textwidth]{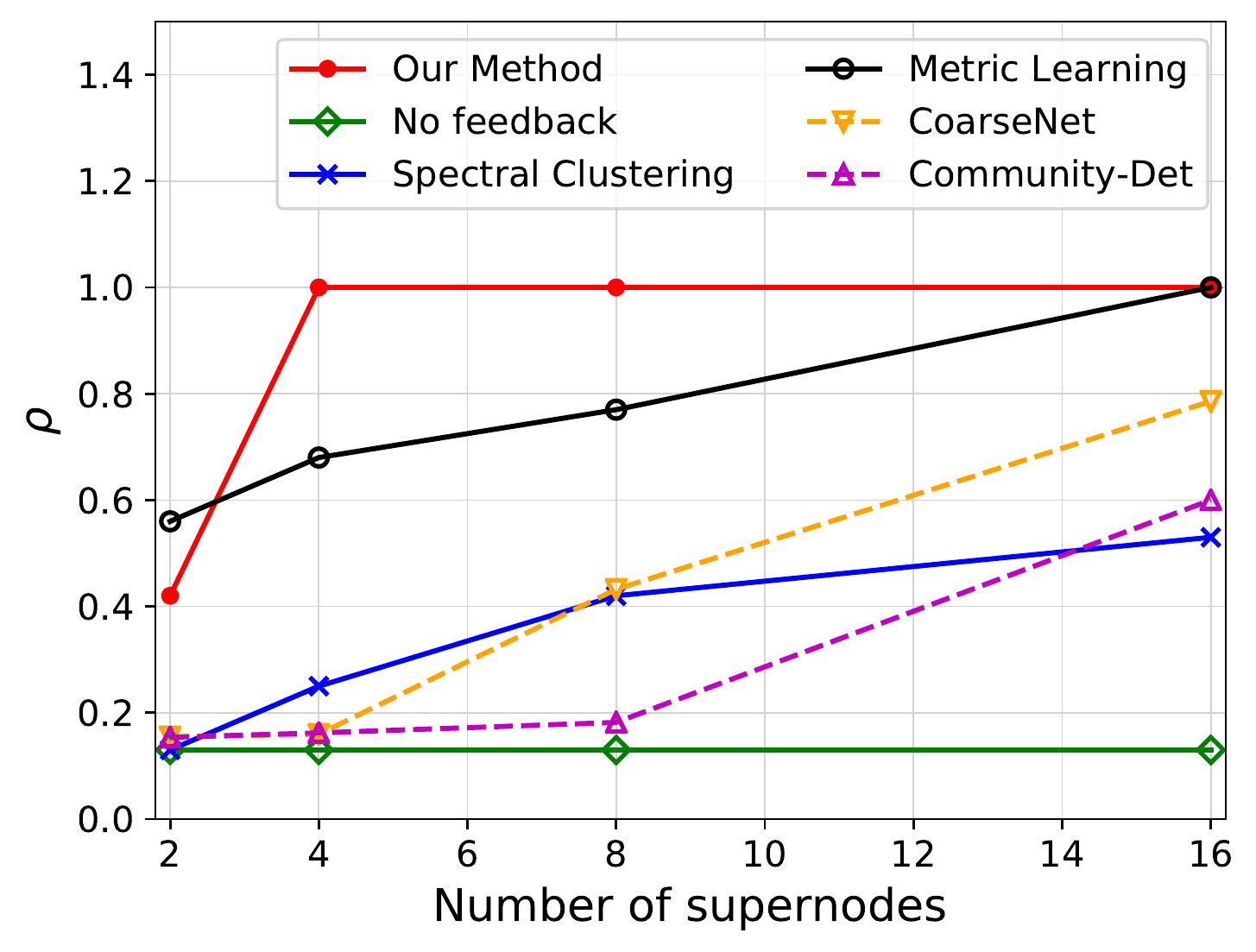}
    \caption{Bert subplot}
    \label{fig:Bert_Takes_a_Powder_VAST2007All_rho}
    \end{subfigure}
    % \begin{subfigure}[htb]{0.23\textwidth}
    % \includegraphics[width =\textwidth]{FIG/Fish_VAST2007All_rho.pdf}
    % \caption{Fish subplot}
    % \label{fig:Fish_VAST2007All_rho}
    % \end{subfigure}
    \begin{subfigure}[htb]{0.23\textwidth}
    \centering
    \includegraphics[width =\textwidth]{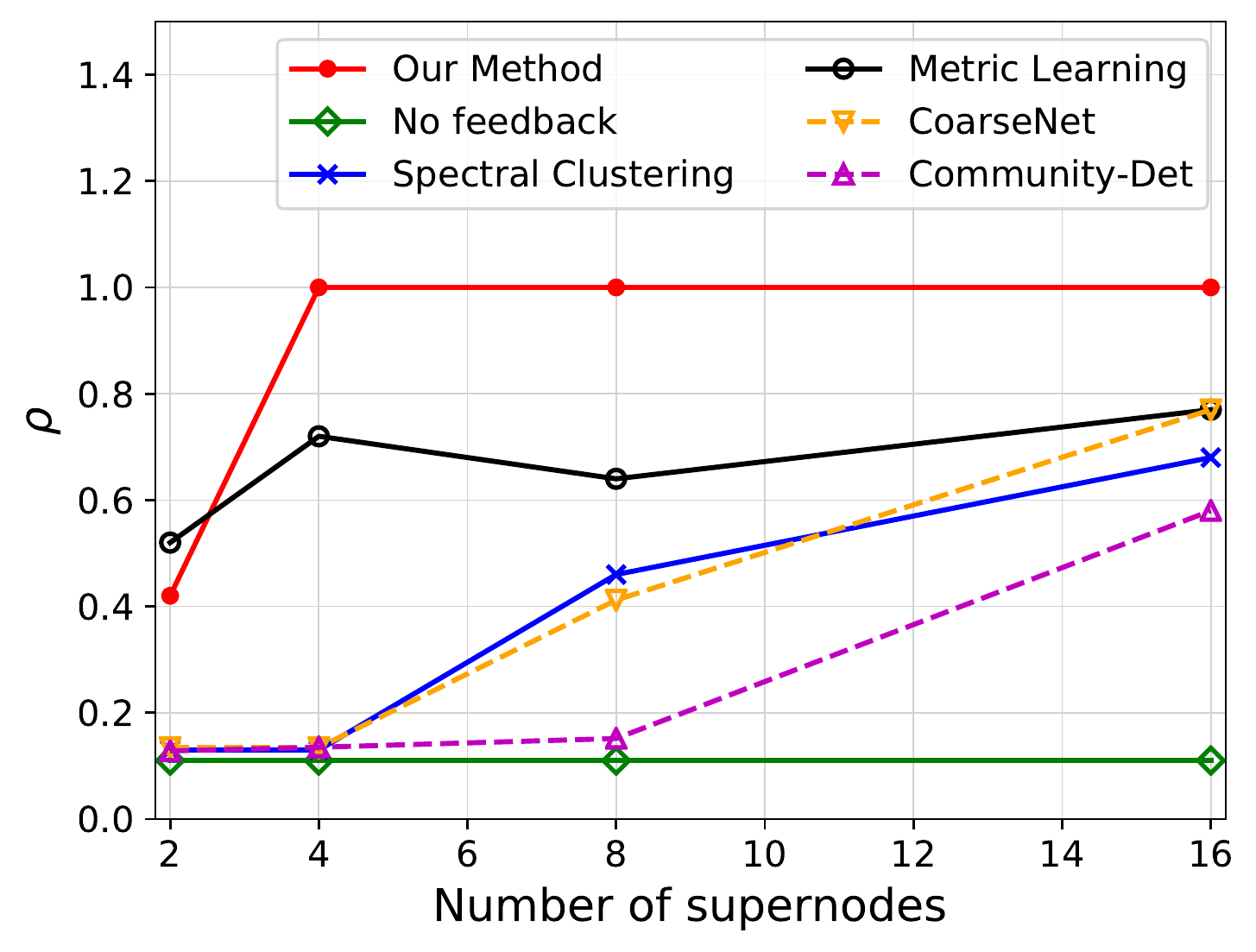}
    \caption{Circus subplot}
    \label{fig:Hassan_Circu_VAST2007All_rho}
    \end{subfigure}
    \caption{Quality of summaries in (a) \crescent and (b-d) different subplots of the \vast dataset. Note, \ourmethod generates network summaries with the highest $\rho$.}
    \label{fig:nof_supernodes_rho}
\end{figure*}

We investigate the quality of summary networks generated by \ourmethod using \crescent and three subplots of \vast datasets with 2, 4, 8, and 16 super-nodes and calculate their $\rho$ values (Eq.~\ref{eq:purity}). In addition, we compare the quality of \ourmethod with baselines.
Figs.~\ref{fig:nof_supernodes_satisfied} and~\ref{fig:nof_supernodes_rho} show the quality of summary networks with various numbers of super-nodes. For each experiment, we randomly selected positive and negative feedback from the ground-truth (see Sec.~\ref{sec:userfeedback}). More specifically, we randomly choose a few pairs of nodes that are relevant to the hidden story as positive feedback and similarly pick pairs in which only one of them is relevant as negative feedback. In all experiments, we fixed the amount of positive and negative feedback at $10\%$ of all possible positive feedback and $1\%$ of all possible negative feedback. %In all figures, we depict positive feedback as solid black lines between documents and  negative feedback as dashed lines.

Fig.~\ref{fig:nof_supernodes_satisfied} shows the ratio of satisfied feedback. The results indicate that \ourmethod generates the highest-quality summary that matches the interests of the user, as it satisfies all of the user's feedback while other baselines can only satisfy part of the feedback. Fig.~\ref{fig:nof_supernodes_rho} shows the quality of summaries $\rho$. \ourmethod generates high-quality summaries (i.e highest $\rho$) networks for various sizes. This implies that users can easily find the relevant documents to scenarios while interacting with visualization generated using \ourmethodviz. It is interesting to mention that because feedback is sparse and the TF-IDF vectors are high-dimensional, the \metric approach is not able to learn proper weights and does not perform well in some cases. Similarly, the poor performance of other graph summarization approaches is explained by the fact that they do not consider user feedback.
% for each term and even in some cases performs worse than \spectral which does not consider any feedback from the user.

\begin{figure*}[t!]
    \centering
    \hideContent{
    \begin{subfigure}[htb]{\textwidth}
    \includegraphics[width = \textwidth]{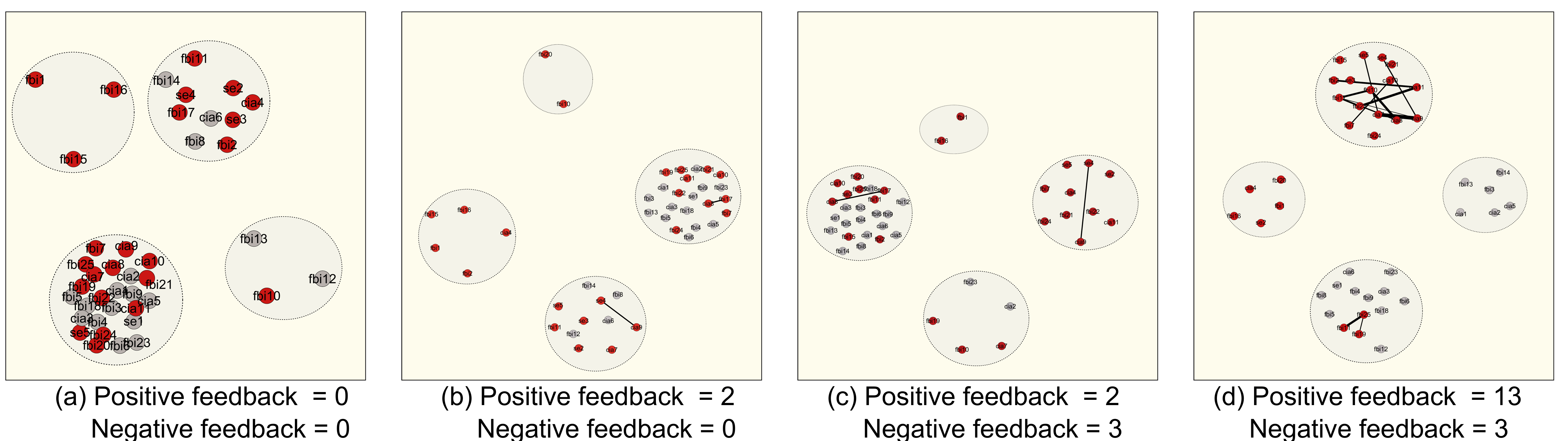}
    \caption{\crescent}
    \label{fig:crescent_evolv}
    \end{subfigure}
    }
    
    \includegraphics[width=\textwidth]{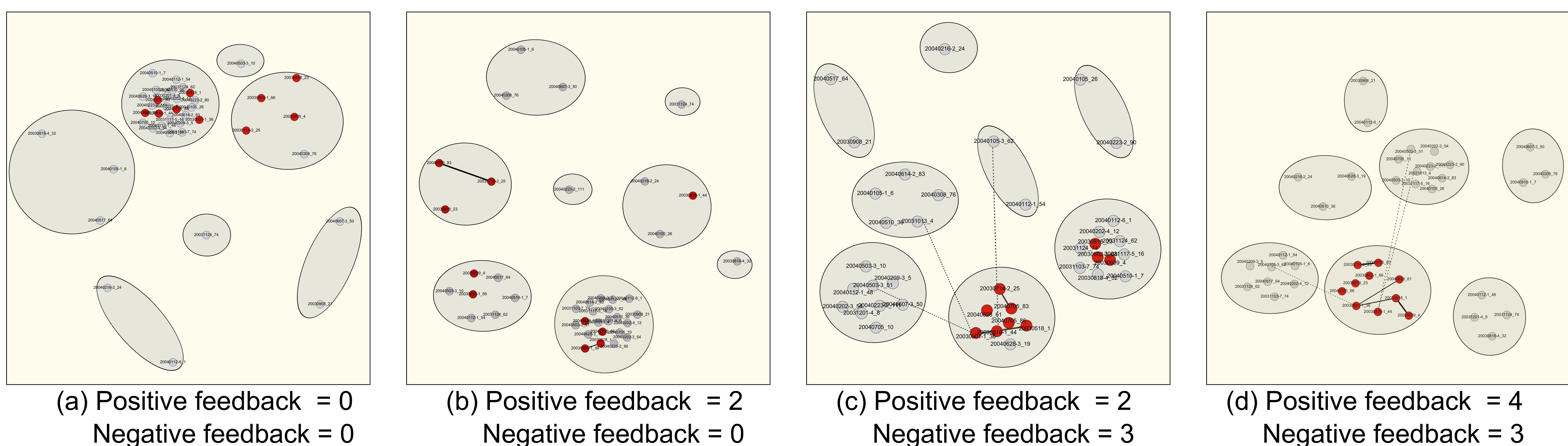}
    %\caption{The network summary evolves with user feedback. (a) Shows the changes in the summary of the \crescent data and (b) shows the summary of the Chinchilla subplot of \vast dataset. Note the black lines represent positive feedback and dashed lines represent negative feedback. Also, red nodes represent relevant documents to the scenario and the gray ones are irrelevant.}
    \caption{The network summary evolves with user feedback. The summary of the Chinchilla subplot of \vast dataset. Note the black lines represent positive feedback and dashed lines represent negative feedback. Also, red nodes represent relevant documents to the scenario and the gray ones are irrelevant.}
    \label{fig:VAST_chinchila_evolve}
\end{figure*}

\hideContent{
\begin{figure*}[htp]
    \centering
    \begin{subfigure}[htb]{0.31\textwidth}
        \includegraphics[height=4.5cm]{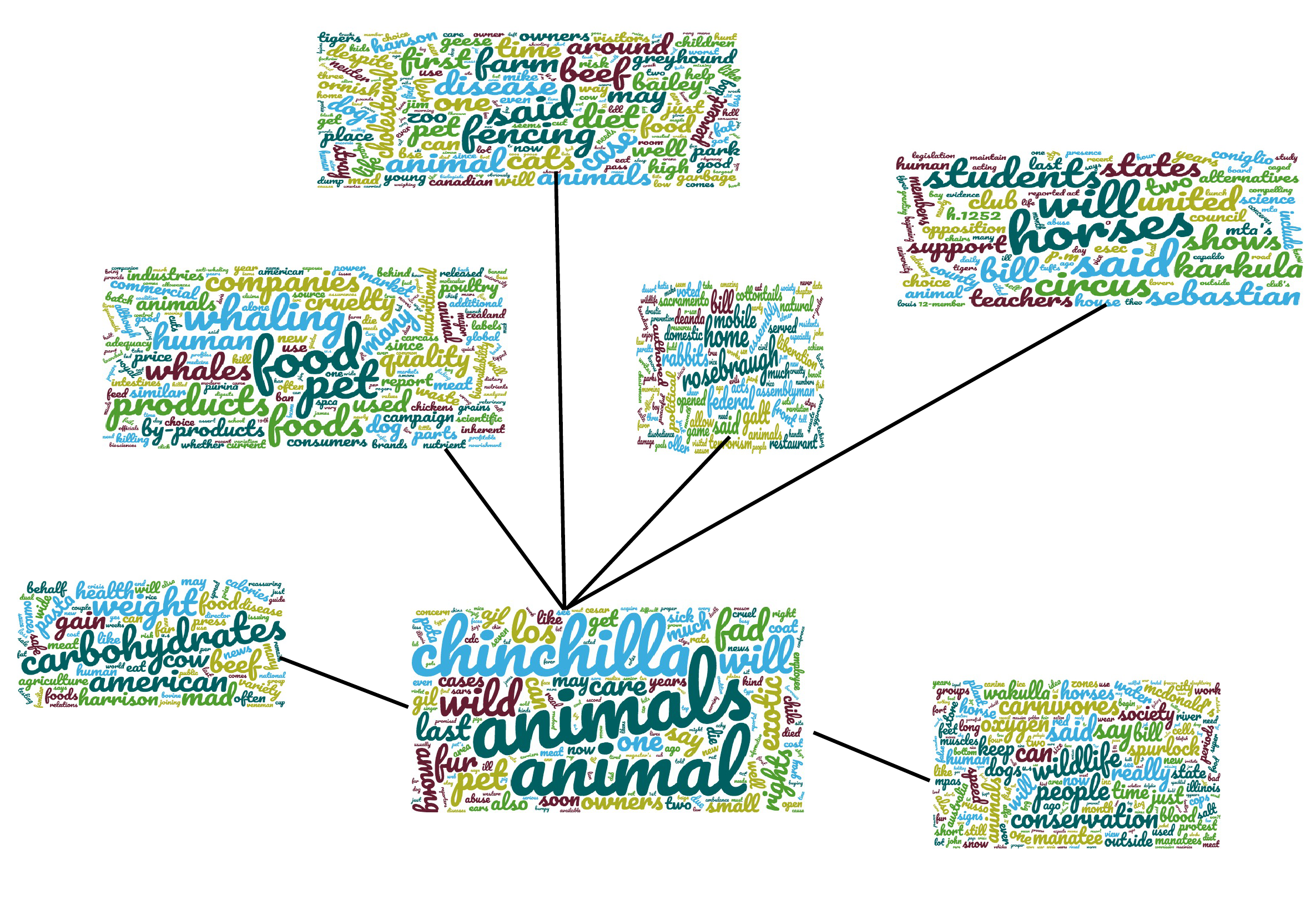}
        \caption{Chinchilla bio-terror: Word-cloud visualization}
        \label{fig:Chinchilla_word_cloud}
    \end{subfigure}
    \begin{subfigure}[htb]{0.35\textwidth}
        \centering
        \includegraphics[height=4.5cm]{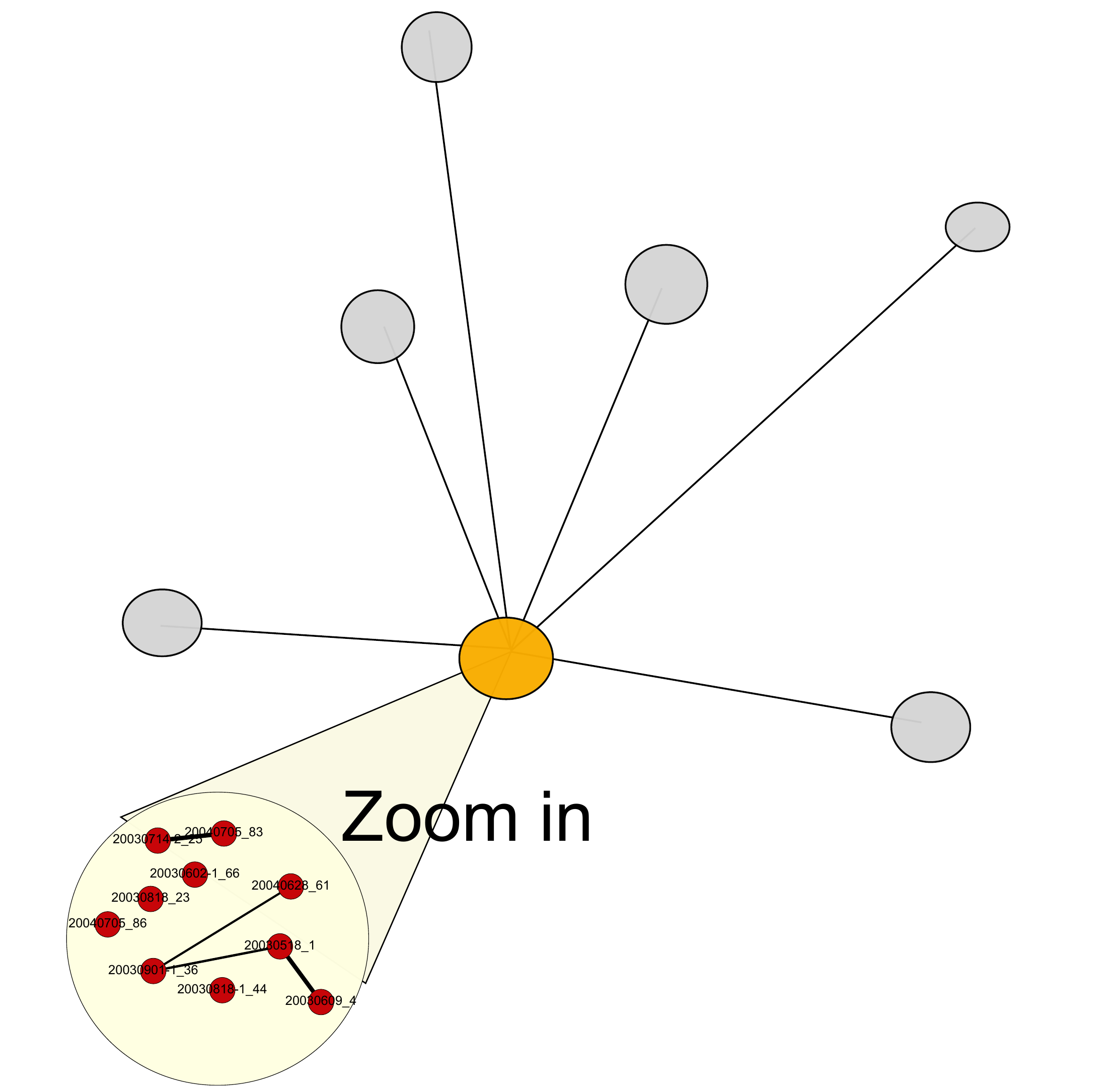}
        \caption{Chinchilla bio-terror: Summary network visualization}
        \label{fig:Chinchilla_unexpanded}
    \end{subfigure}
    \begin{subfigure}[htb]{0.33\textwidth}
        \includegraphics[height=4.5cm]{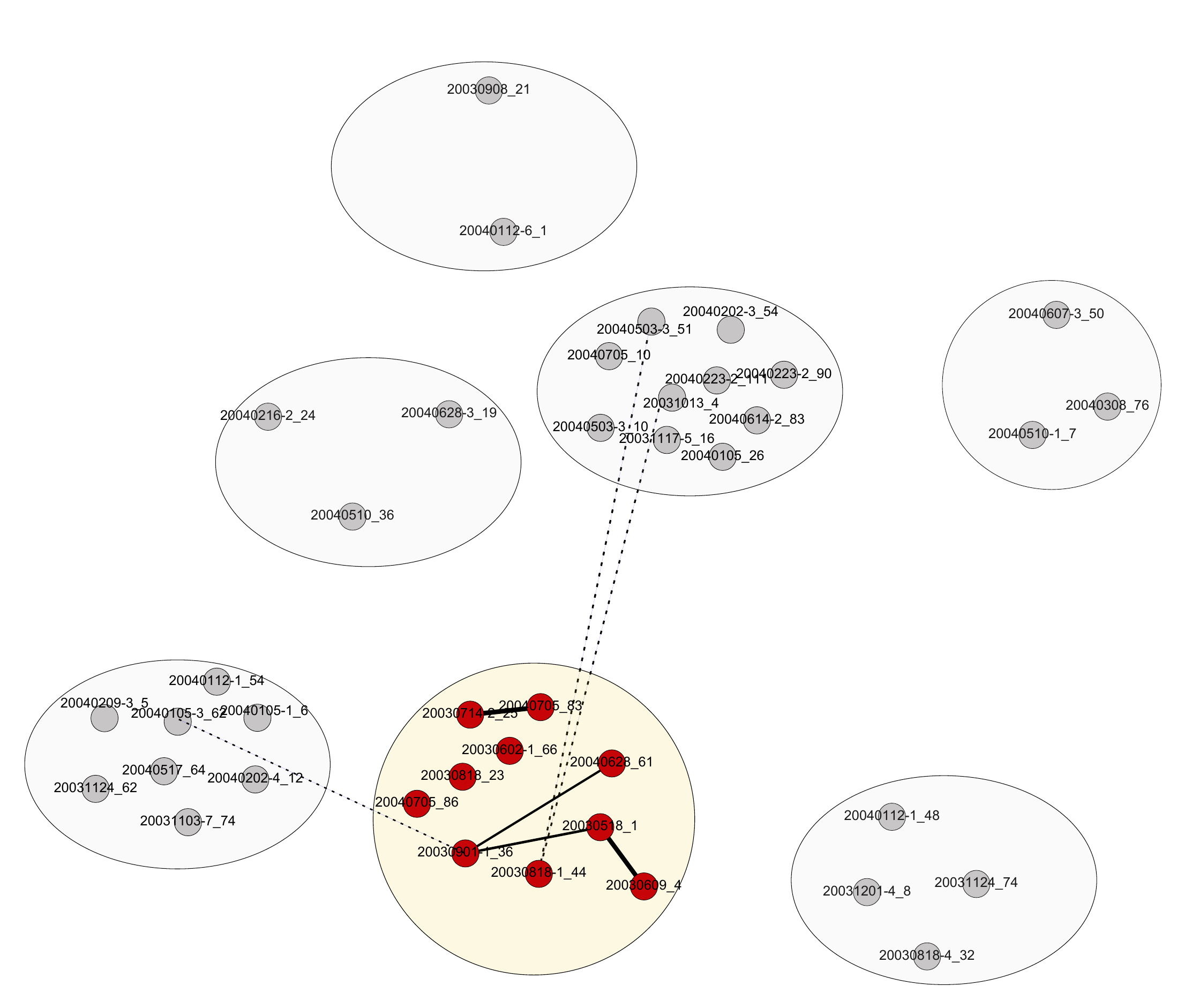}
        \caption{Chinchilla bio-terror: Expanded visualization}
        \label{fig:Chinchilla_expanded}
    \end{subfigure}

    %\vspace{-0.2cm}
    \caption{The visualization of the Chinchilla bio-terror (a) shows the word cloud of the documents of each super-node in the summary network. (b) shows the visualization of the summary graph. \ourmethod expands the super-node that the user selects and shows its documents for further investigation. (c) show the expanded visualization of all super-nodes. Note, red nodes represent relevant documents to the scenario and the gray ones are irrelevant.}
    \label{fig:vast_vis}
\end{figure*}
}

\subsection{Q2. Effect of Feedback}
\label{sec:feedback}
We investigate how \ourmethod evolves the summary of a document network while the user gives positive and negative feedback. We also objectively measure the change in quality of super-nodes based on the feedback by tracking the changes in $\rho$. Similar to Sec.~\ref{sec:quality_summary}, feedback is randomly generated from the ground-truth.

To showcase the quality of \ourmethod on the \vast dataset, for each subplot we extract a subset of documents relevant to the subplot.  Fig. \ref{fig:VAST_chinchila_evolve} shows the visualization of the Chinchilla Bio-terror subplot.  We depict positive feedback as solid black lines between documents and  negative feedback as dashed lines.  Initially our method can only identify four of the related documents with the subplot and puts them in a super-node. However, the rest of the relevant documents are mixed with other irrelevant ones in the largest super-node of the summary network. Next, the user gives feedback regarding the similarity of two pairs of documents. \ourmethod updates the visualization (Fig.~\ref{fig:VAST_chinchila_evolve}(b)). However, this is not enough to improve the quality. When the user adds the negative feedback, \ourmethod can distinguish more relevant documents (Fig.~\ref{fig:VAST_chinchila_evolve}(c)). Finally, by giving two more positive feedback interactions, \ourmethod can accurately identify the relevant documents with the subplots and puts them in a separate super-node (Fig.~\ref{fig:VAST_chinchila_evolve}(d)).

%% file: 070conclusion.tex
In this paper, we explored the problem of learning interactive network summaries with an application of generating multi-level and generalizable visualization models for text analysis. We proposed a novel and effective network summarization algorithm, \ourmethod, which leverages a feedback-based reinforcement learning approach to incorporate human input. We also proposed \ourmethodviz as a framework to produce a visualization based on hierarchical network summaries generated by \ourmethod. Our experiments show that \ourmethod is able to summarize and \ourmethodviz is able visualize a document network meaningfully to reveal hidden stories in the corpus and connect the dots between documents.

As \ourmethod relies on Q-learning, it can be made faster, which is also a promising direction for future work. As shown by our experiments, it already works well on real document networks and solves real tasks in practice. In the future, we plan to apply this interactive network summarization model to much larger document datasets and temporal data scenarios. Moreover, the flexibility we obtain from the reinforcement learning approach makes it possible to bring learning into summarization and enable better generalization and personalization.  For example, we can build a personalized interactive summarization model for each user to reflect their interests and quickly summarize different datasets without requiring user input for each new corpus.

Network summarization can lead to other meaningful visualizations by 
incorporating more diverse semantic interactions into the reinforcement learning approach. For example, we would like to explore how to differentiate between highlighting, overlapping, and annotating documents in our framework. Also, leveraging more visual encodings to create a more understandable and user-friendly summarization is a  fruitful direction. We can explore using our approach for summarizing and visualizing other data types such as social networks and images as well. Our approach here  opens several additional interesting avenues for future work.

%% file: main.bbl
\begin{thebibliography}{10}

\bibitem{adhikari2018propagation}
B.~Adhikari, Y.~Zhang, S.~E. Amiri, A.~Bharadwaj, and B.~A. Prakash.
\newblock Propagation-based temporal network summarization.
\newblock {\em IEEE Transactions on Knowledge and Data Engineering},
  30(4):729--742, 2018.

\bibitem{amiri2018efficiently}
S.~E. Amiri, L.~Chen, and B.~A. Prakash.
\newblock Efficiently summarizing attributed diffusion networks.
\newblock {\em Data Mining and Knowledge Discovery}, 32(5):1251--1274, 2018.

\bibitem{bradel2014multi}
L.~Bradel, C.~North, L.~House, and S.~Leman.
\newblock Multi-model semantic interaction for text analytics.
\newblock In {\em 2014 {IEEE} Conference on Visual Analytics Science and
  Technology, {VAST} 2014, Paris, France, October 25-31, 2014}, pages 163--172,
  2014.

\bibitem{dwyer2005dig}
T.~Dwyer and Y.~Koren.
\newblock Dig-cola: directed graph layout through constrained energy
  minimization.
\newblock In {\em INFOVIS}. IEEE, 2005.

\bibitem{DBLP:journals/tvcg/EndertFN12}
A.~Endert, P.~Fiaux, and C.~North.
\newblock Semantic interaction for sensemaking: Inferring analytical reasoning
  for model steering.
\newblock {\em {IEEE} Trans. Vis. Comput. Graph.}, 18(12):2879--2888, 2012.

\bibitem{girvan2002community}
M.~Girvan and M.~E. Newman.
\newblock Community structure in social and biological networks.
\newblock {\em Proceedings of the national academy of sciences},
  99(12):7821--7826, 2002.

\bibitem{grinstein2007vast}
G.~Grinstein, C.~Plaisant, S.~Laskowski, T.~O'Connell, J.~Scholtz, and
  M.~Whiting.
\newblock Vast 2007 contest-blue iguanodon.
\newblock In {\em 2007 IEEE Symposium on Visual Analytics Science and
  Technology}, pages 231--232. IEEE, 2007.

\bibitem{hughes2003discovery}
F.~Hughes and D.~Schum.
\newblock Discovery-proof-choice, the art and science of the process of
  intelligence analysis-preparing for the future of intelligence analysis.
\newblock {\em Joint Military Intelligence College, Washington, DC.}, 2003.

\bibitem{karypis:metis:sc98}
G.~Karypis and V.~Kumar.
\newblock Multilevel algorithms for multi-constraint graph partitioning.
\newblock In {\em Supercomputing, 1998. SC98. IEEE/ACM Conference on}, pages
  28--28. IEEE, 1998.

\bibitem{purohit2014fast}
M.~Purohit, B.~A. Prakash, C.~Kang, Y.~Zhang, and V.~Subrahmanian.
\newblock Fast influence-based coarsening for large networks.
\newblock In {\em Proceedings of the 20th ACM SIGKDD international conference
  on Knowledge discovery and data mining}, pages 1296--1305. ACM, 2014.

\bibitem{shah2015timecrunch}
N.~Shah, D.~Koutra, T.~Zou, B.~Gallagher, and C.~Faloutsos.
\newblock Timecrunch: Interpretable dynamic graph summarization.
\newblock In {\em Proceedings of the 21th ACM SIGKDD International Conference
  on Knowledge Discovery and Data Mining}, pages 1055--1064. ACM, 2015.

\bibitem{shipman1999formality}
F.~M. Shipman and C.~C. Marshall.
\newblock Formality considered harmful: Experiences, emerging themes, and
  directions on the use of formal representations in interactive systems.
\newblock {\em Computer Supported Cooperative Work (CSCW)}, 8(4):333--352,
  1999.

\bibitem{sutton1998introduction}
R.~S. Sutton and A.~G. Barto.
\newblock {\em Introduction to reinforcement learning}, volume 135.
\newblock MIT press Cambridge, 1998.

\bibitem{von2007tutorial}
U.~Von~Luxburg.
\newblock A tutorial on spectral clustering.
\newblock {\em Statistics and computing}, 17(4):395--416, 2007.

\bibitem{watkins1992q}
C.~J. Watkins and P.~Dayan.
\newblock Q-learning.
\newblock {\em Machine learning}, 8(3-4):279--292, 1992.

\bibitem{DBLP:conf/sdm/WhangDG15}
J.~J. Whang, I.~S. Dhillon, and D.~F. Gleich.
\newblock Non-exhaustive, overlapping \emph{k}-means.
\newblock In {\em Proceedings of the 2015 {SIAM} International Conference on
  Data Mining, Vancouver, BC, Canada, April 30 - May 2, 2015}, pages 936--944,
  2015.

\bibitem{xing2003distance}
E.~P. Xing, M.~I. Jordan, S.~J. Russell, and A.~Y. Ng.
\newblock Distance metric learning with application to clustering with
  side-information.
\newblock In {\em Advances in neural information processing systems}, pages
  521--528, 2003.

\end{thebibliography}
